\documentclass[10pt, a4paper]{article}
\usepackage{lrec}
\usepackage{multibib}
\newcites{languageresource}{Language Resources}
\usepackage{graphicx}
\usepackage{tabularx}
\usepackage{soul}
\usepackage{epstopdf}
\usepackage{booktabs}
\usepackage{arydshln}
\usepackage[latin1]{inputenc}
\usepackage{url}
\usepackage{xstring}
\usepackage{arabtex}
\usepackage{multirow}

\title{ArSentD-LEV: A Multi-Topic Corpus for Target-based Sentiment Analysis in Arabic Levantine Tweets}

\name{Ramy Baly$^{(1)}$, Alaa Khaddaj$^{(2)}$, Hazem Hajj$^{(2)}$, Wassim El-Hajj$^{(3)}$, Khaled Bashir Shaban$^{(4)}$}
\address{
(1) MIT Computer Science and Artificial Intelligence Laboratory, Cambridge, MA 02139, USA\\
(2) American University of Beirut, Electrical and Computer Engineering Department, Beirut, Lebanon\\
(3) American University of Beirut, Computer Science Department, Beirut, Lebanon\\
(4) Qatar University, Computer Science and Engineering Department, Doha, Qatar \\
\url{baly@mit.edu}, \url{{awk11,hh63}@aub.edu.lb}, \url{we07@aub.edu.lb}, \url{khaled.shaban@qu.edu.qa}}

\abstract{
Sentiment analysis is a highly subjective and challenging task. Its complexity further increases when applied to the Arabic language, mainly because of the large variety of dialects that are unstandardized and widely used in the Web, especially in social media.
While many datasets have been released to train sentiment classifiers in Arabic, most of these datasets contain shallow annotation, only marking the sentiment of the text unit, as a word, a sentence or a document.
In this paper, we present the Arabic Sentiment Twitter Dataset for the Levantine dialect (ArSenTD-LEV).
Based on findings from analyzing tweets from the Levant region, we created a dataset of 4,000 tweets with the following annotations: the overall sentiment of the tweet, the target to which the sentiment was expressed, how the sentiment was expressed, and the topic of the tweet.
Results confirm the importance of these annotations at improving the performance of a baseline sentiment classifier.
They also confirm the gap of training in a certain domain, and testing in another domain.\\
}

\begin{document}
\maketitleabstract

\section{Introduction}
\label{sec:introduction}
Sentiment analysis refers to the task of inferring opinions from text~\cite{liu2012sentiment}.
Research in sentiment analysis has been driven by the interest in its wide range of applications and the availability of large amounts of subjective data on the Web~\cite{ravi2015survey}.
Today's social media has provided people the opportunity to connect across the globe and express their opinions and emotions freely and abundantly. Twitter is one of the most used social media platforms, with recent statistics~\footnote{\url{https://www.socialbakers.com/statistics/twitter/}} indicating that over 500 million tweets are being sent out daily, mainly to express opinions about personal or trending topics, news or events~\cite{sareah2015interesting}.

Sentiment analysis has been widely approached as a text classification problem, with the target of predicting the overall opinion of a given text (words, sentences or documents)~\cite{pang2002thumbs,socher2013recursive,tang2015document,farra2010sentence}.
However, sentiment analysis can also be performed at more granular levels, such as identifying target entities~\cite{brody2010unsupervised,somasundaran2009recognizing,farra2017smarties} and predicting opinions towards these targets, whether in Twitter~\cite{jiang2011target}, online comments~\cite{biyani2015entity} or product reviews~\cite{wang2016attention,kirange2015emotion}.
These tasks are critically-important to handle cases where the text contains multiple opinions expressed towards one or different targets, which is a common phenomenon in product reviews.

Research in exploring methods for English sentiment analysis has been leading the way, while other languages, including Arabic, still lag behind. Most advances were made in English, mainly because to the availability of sentiment corpora to support such tasks. This paper aims at providing new resources to support research advances in Arabic. As a matter of fact, Arabic ranks as the 4\textsuperscript{th} most spoken language worldwide~\cite{paolillo2006evaluating}, and as of March 2017, 11.1 million Twitter users from the Arab world are generating 27.4 million tweets on a daily basis~\cite{salem2017social}.

In the last few years, there has been a significant progress in creating resources for Arabic sentiment analysis. However, these resources are often coupled with sentiment annotations only, and typically on a  three point scale (1 to 3) instead of the common 5-point typically used in reviews, which also reflects sentiment intensity. Furthermore, it was found that modeling sentiment depends on the domain or topic at hand, and that a sentiment model trained on one domain is not expected to perform as well on another~\cite{pan2010cross}.
Additionally, textual semantics vary across languages and dialects~\cite{baly2017characterization} due to cultural factors~\cite{salameh2015sentiment,mohammad2016translation}.
For example, \<sb.hAn Al-lah Al`ly Al`.zym> {\it Glory to God the Great} is used in the Levant to express positive sentiment, whereas it is considered a religious saying with no sentiment in other Arab regions, {\it e.g.} the Gulf countries.
Consequently, cross-lingual and cross-domain approaches~\cite{chen2016adversarial,li2017end} have been explored to avoid the need for a sentiment corpus for each domain or language, which is costly and time-consuming.

In this paper, we address the limitations of having a corpus annotated for sentiment only, by creating a corpus and having it simultaneously annotated for different and important aspects needed for research in sentiment analysis.
We create our corpus from Twitter content due to its widespread use in the Arab world.
Given the cultural and linguistic differences across Arab regions, causing shifts in semantics, we focus on developing sentiment models for the Levantine dialect.
According to~\cite{zaidan2014arabic}, Arabic dialects can be categorized into Egyptian, Levantine, Gulf, Iraqi and Maghrebi. Our corpus is composed of tweets retrieved from Levantine countries (Jordan, Lebanon, Palestine and Syria), where the Levantine dialect is the 3\textsuperscript{rd} most spoken Arabic dialect~\cite{zaidan2014arabic}. We selected a group of 4,000 tweets, and had users annotate those tweets via crowdsourcing to: 1) identify the sentiment targets in each tweet, 2) annotate both sentiment polarity and intensity on a five-point scale, from {\it very negative} to {\it very positive}, 3) indicate whether sentiment was expressed {\it implicitly} or {\it explicitly}, and 4) finally to identify the topic the tweet is discussing.
This corpus is publicly available.~\footnote{The corpus is available at \url{www.oma-project.com}}

The resulting corpus provides a resource complement to existing Arabic dialect resources~\cite{baly2017comparative,assiri2016saudi,refaee2014arabic}.
It will also enable models that can exploit sentiment target identification, topic identification and sentiment expression. Furthermore, it will open doors to investigate cross-dialect sentiment models by leveraging existing Twitter corpora from other regions and dialects.
Several experiments are conducted to confirm the benefits of such new aspects~\cite{joty2017cross}.
We show that topic-based models outperform models that do not consider the topic of the text. 

The remaining of the paper is organized as follows.
Section~2 describes previous efforts to create sentiment datasets in Arabic.
Section~3 presents an analysis of Arabic tweets and describes our methodology to create and annotate the corpus.
Section~4 presents experimental results to benchmark the performance of a baseline classifier on our developed corpus, and also to emphasize the impact of topic change on the performance.
Concluding remarks are made in Section~5.

\section{Related Work}
\label{sec:related_work}


Sentiment analysis has been performed by training machine learning models using different choices of features~\cite{abdul2011subjectivity,abdul2014samar,badaro2014large,refaee2014subjectivity,badaro2015light,al2015deep,baly2016meta,baly2017omam,al2017aroma}.
However, training and evaluating accurate sentiment models requires the availability of corpora with sentiment labeling.
Below, we list commonly-known Arabic sentiment corpora.

\newcite{abdul2011subjectivity} created a corpus by annotating 2,855 sentences, coming from the first 400 documents of the Penn Arabic Treebank Version 1 Part 3~\cite{maamouri2004penn}, using the following labels: objective, subjective-positive, subjective-negative and subjective-neutral.
This dataset was extended by annotating additional 5,342 sentences from Wikipedia talk pages and 2,532 sentences from web forums to create the AWATIF corpus~\cite{abdul2012awatif}.
\newcite{rushdi2011oca} created the Opinion Corpus for Arabic (OCA), which consists of 500 Arabic movie reviews that are annotated as either positive or negative.
\newcite{aly2013labr} created LABR; a large-scale corpus consisting of 63,257 book reviews written in Arabic, each rated on a five-point scale.
\newcite{elsahar2015building} retrieved 33,116 Arabic reviews on movies, hotels, restaurants and products, and automatically annotated them using available ratings.

The above-mentioned corpora contained data written in Modern Standard Arabic (MSA).
Additional efforts have been made to develop corpora for dialectal Arabic, due to its widespread use in the Web.
\newcite{refaee2014arabic} retrieved 8,868 tweets from multiple Arabic dialects, and annotated them for both subjectivity and sentiment using the following labels: polar, positive, negative, neutral and mixed.
\newcite{baly2017sentiment} created the Arabic Sentiment TreeBank (ArSenTB) using 1,176 comments, from the QALB dataset~\cite{mohit2014first}, written in MSA and a mixture of different dialects. In addition to sentence-level sentiment annotation, comments were transformed into phrase structure parse trees, and the sentiment of each constituent (node in the tree) was also annotated, totaling up to 123,000 constituents.
\newcite{al2016prototype} created a corpus covering MSA as well as several Arabic dialects. This corpus is composed of 1,442 reviews extracted from five domains: economy, food-life style, religion, sports and technology. Annotation was performed manually to ensure high quality.
\newcite{nabil2015astd} created the Arabic Sentiment Tweets Dataset (ASTD), which consists of 10,006 tweets, written in the Egyptian dialect and annotated as positive (799), negative (1,684), mixed (832) or objective (6,691).
\newcite{medhaffar2017sentiment} created the Tunisian Sentiment Analysis Corpus (TSAC) by retrieving 17,000 comments written with Tunisian dialect from Facebook, and annotating them as positive or negative.
\newcite{baly2017characterization} created two datasets, each consisting of 1,000 tweets, written in Egyptian and Emarati dialects and manually annotated for sentiment at a 5-point scale, from very negative to very positive.
A similar effort was done to create AraSenti-Tweet; a sentiment corpus of 17,573 tweets written in MSA and in Saudi dialect~\cite{al2017arasenti}.

It can be observed that, despite the recent efforts to create Arabic sentiment corpora, the majority of these datasets only focused on labeling the overall sentiment of the text, while ignoring other useful information, such as the target of the sentiment and the topic being discussed.
A corpus with similar annotations was developed for SemEval-2016 Task~4 on Sentiment Analysis in Arabic tweets~\cite{rosenthal2017semeval}. The corpus consisted of 3,355 tweets annotated  by the polarity of sentiment in the tweet and the sentiment towards a specific target in the tweet (also known as stance).
Also,~\cite{al2015human} used a subset of 2,800 reviews from the LABR corpus and enriched it with aspect-based sentiment annotations.

In this paper, we present {\sc ArSenTD-LEV}; an Arabic sentiment dataset that is composed of Levantine tweets, and we enrich it with a variety of sentiment-related annotations that never existed together in a single corpus.
\section{Dataset}
\label{sec:dataset}

In this section, we describe our methodology to create the new sentiment corpus.

\subsection{Manual Data Analysis}
To have the proper guidelines in the annotation process, we conducted manual analysis to make sure we have solid insights into the intricacies of the sentiment analysis and the required sentiment annotations.
The goal of the analysis was to gain insights and understand the characteristics and different usages of Twitter in the Levant region.
As such, a sample of 200 tweets, generated in countries from the Levant region, were retrieved and characterized.
We focused on information that should be critical to developing accurate sentiment analysis models, including: the topic being discussed, the language being used, the way sentiment was being expressed and the target of the sentiment.

\paragraph{Topic Analysis}
The first question we wanted to answer is: {\it what topics are often discussed on Twitter?}.
Our findings, shown in Table~\ref{tab:topics_pilot}, suggest that most of the tweets expressed opinions about personal and daily matters, and to a less extent on political issues, especially the ongoing conflicts in the Middle East.
People also discussed religious matters and tend to quote verses from the Quran.
Table~\ref{tab:topics_pilot} also illustrates the different items discussed per topic, ordered from most to least frequent in the sample set.
In addition to the outcome of knowing which topics were being discussed, we also used the sample tweets to identify the most discriminative keywords across topics, which are used later when creating the corpus.

\begin{table}[h!]
\centering
\begin{tabular}{llm{1.75in}}
\toprule
{\bf Topic} & {\bf Size } & {\bf Sub-topics}\\ \midrule
Personal & 36\% & sarcasm, love, sadness and optimism \\ \midrule
Politics & 23\% & Syrian war, Palestinian war, Lebanese elections, revolution and terrorism \\ \midrule
Religion & 11\% & sermon, mention, praising God, religious events and Quranic verses\\ \midrule
Sports & 6\% & international and local soccer games, soccer players and basketball\\ \midrule
Other & 24\% & entertainment, ads, health, education, economy, technology and weather\\
\bottomrule
\end{tabular}
\caption{Breakdown of the different topics and sub-topics that were discussed in the sample set of 200 tweets.}\label{tab:topics_pilot}
\end{table}

\paragraph{Language Use}
By analyzing the language that was used to write the 200 tweets, we observed that:
51\% were written in Modern Standard Arabic (MSA),
34\% in Levantine dialect, and
the remaining 15\% in English, Arabizi, or a mixture of MSA and dialectal Arabic (DA).
We also observed that most personal tweets were written in DA, indicating that users prefer to use it rather than MSA when it comes to discussing personal aspects of their lives and feelings.

\paragraph{Sentiment Expression}
We analyzed the sentiment distribution in the 200 tweets by labeling the sentiment polarity and the way it was expressed, i.e., explicitly or implicitly, for each tweet.
We observed that a significant amount of tweets were negative, which can be attributed to the current political situation having a direct impact on people's lives and opinions.
We also observed that sentiment distribution changes from one country to another; it is mostly negative in Syria and mostly neutral in Jordan, which may reflects the countries' political and social stabilities.
Finally, among the subjective tweets, sentiment was expressed explicitly in 64\% and implicitly in 35\% of the tweets, which is an indication of the complexity in opinion mining.

\subsection{Corpus Development}
Our goal is to create an Arabic dataset of tweets from the Levant region, and annotate them for topic, sentiment polarity, sentiment intensity, sentiment target and sentiment expression.
In order to create this corpus we performed the following steps.

\paragraph{Tweets Retrieval}
We used the {\sc TweePy} python module to retrieve tweets using pre-specified geo-locations covering four countries from the Levant region: Jordan, Lebanon, Palestine and Syria. The retrieval process began on November 5\textsuperscript{th} 2017 and ended on November 29\textsuperscript{th} 2017.
As a result, we retrieved 45,000 tweets that are equally distributed across the four countries.

\paragraph{Pre-processing}
The target size of our corpus is 4,000 tweets; 1,000 for each country.
We also aim to collect tweets discussing the common topics (politics, religion, sports, personal and entertainment) that we encountered in the manual analysis.
Therefore, we created for each of the five topics a list of topic-specific keywords; for each topic we selected the most frequent words in the sample set that were the most discriminative with regard to that topic.
We checked the 45K tweets against these lists and kept those that contained at least one keyword from one list and none from the others.
This is a naive topic classification that will not be part of the final corpus, and that was performed only to increase the likelihood of having tweets discussing our target topics.
We also excluded tweets written in foreign languages and those only containing URLs and emoticons.
Finally, for each country, we selected the longest 1,000 tweets such that they are balanced across our target topics.
It is worth mentioning that despite the fact that we enforced some balance over the different topics, we do not expect this to be the case in the final corpus after manual annotation, since topics are inherently imbalanced as shown in Table~\ref{tab:topics_pilot}.

\begin{table*}[h!]
\centering
\begin{tabular}{l:c|l:c|l:c}
\toprule
\multicolumn{2}{l}{\bf Topic} & \multicolumn{2}{|l}{\bf Sentiment} & \multicolumn{2}{|l}{\bf Expression} \\ \midrule
Personal & 32.6\% & Very negative & 16.3\% & Explicit & 73.6\%\\
Sports & 12.12\% & Negative & 30.8\% & Implicit & 4.3\% \\
Politics & 37.63\% & Neutral & 22.13\% & None & 22.1\% \\
Religions & 9.83\% & Positive & 20.1\% & &  \\
Entertainment & 4.35\% & Very positive & 10.7\% & & \\
Other & 3.45\% & & & & \\
\bottomrule
\end{tabular}
\caption{Distributions of the different annotated features in the corpus.}\label{tab:stats}
\end{table*}

\paragraph{Annotation}
The annotation process was carried out via crowdsourcing and using the CrowdFlower platform.
For each tweet, annotators were instructed to 1) select its overall sentiment, 2) identify the target of this sentiment in the tweet (in case it was not neutral) by copying segments of the tweet into a text box, 3) identify whether the sentiment was expressed explicitly or implicitly, and 4) specify the topic being discussed.
Sentiment labels were assigned based on a 5-point scale using the following labels: {\it very negative}, {\it negative}, {\it neutral}, {\it positive} and {\it very positive}.
Motivated by our manual analysis of a sample of tweets, we pre-defined the following topics: {\it politics}, {\it religions}, {\it sports} and {\it personal}. If a tweet's topic did not belong to one of these choices, annotators will have to specify another topic based on their own judgment.
Before conducting the large-scale annotation task, we conducted a pilot task to ensure the clarity of the guidelines and examples, and consequently the task.

Tweets were randomly assigned to at least 5 annotators, and up to 4 additional annotators were asked to participate in case of ties.
As a result, we had 5-9 different annotators annotating each tweet, which is a reasonable number to perform aggregation over 5 classes.
Annotations were aggregated based on majority voting, and the annotators' trust score (reflecting their work accuracy) was used for breaking ties.
To make sure only qualified annotators are allowed to do the task, we performed quality control by creating a gold set of 181 tweets that we annotated for sentiment, and used it to monitor the annotators' accuracy on this set.
Only those with an accuracy higher than 75\% were allowed to stay on the job.

\paragraph{Post-Processing}
To aggregate sentiment targets returned by annotators, we automatically extracted the longest common substring among targets whose annotators agreed with the final aggregated label.
In other words, if the aggregated sentiment was {\it positive}, we only considered the pool of targets returned by annotators who annotated the tweet as either {\it positive} or {\it very positive}.
Also, while we instructed annotators that the sentiment target must be explicitly observed in the tweet, we observed that in 160 tweets, annotators specified the targets with their own wording. We resolved these cases manually.
We also manually aggregated the topic annotations of 138 tweets whose topic was not one of the pre-specified topics.

\subsection{Statistics and Evaluation}
It is of critical importance to evaluate the annotation quality to make sure the corpus can be properly used to develop accurate sentiment models.
We evaluated how well annotators of the each tweet agreed on the same label.
Over a sample of 100 tweets, the average agreement was
83\% for topics,
73\% for sentiments, and
72\% for sentiment expressions.
These numbers are significantly higher than 50\% (the case of a tie), indicating a straightforward majority-based aggregation for most of the tweets.
Differences in agreements reflect the relative difficulty of the task. For instance, it can be inferred that identifying the sentiment of a tweet and how it was expressed is a more ambiguous and subjective task than identifying the topic.
It is worth mentioning that the agreement on sentiment increases up to 81\% when considering three sentiment classes, which indicates that many cases of disagreement were due to differences in annotating the intensity.

We also report a 83\% agreement between the labels of the gold set (181) tweets, and the aggregation of the CrowdFlower-annotated sentiments for the same tweets.
In order to evaluate the quality of sentiment targets, we manually annotated the targets for the gold set of tweets, and compared them to the outcome of selecting the longest common substring among CrowdFlower-annotated targets for the same tweets.
By counting the number of common words between both targets and normalizing it by the length of the gold target, we found a 63\% overlap, on average, which is acceptable given the highly-subjective nature of the task.
Finally, statistics and distributions of the different annotated features from the corpus are presented in Table~\ref{tab:stats}.
\section{Experiments and Results}
\label{sec:experiments_results}
In this section, we present the results of applying a baseline sentiment classifier on our new corpus: ArSenTD-LEV.
We also perform cross-topic and in-topic experiments to emphasize the impact of changing the topic between training and testing data, and also by using the {\it topic} and {\it sentiment expression} as additional features to train the classifier.

Our feature set is composed of {\it uni}-grams and {\it bi}-grams represented with TF-iDF scores.
These features were used to train different classifiers including logistic regression, Support Vector Machines (SVM), random forest trees and the ridge classifier.
We report only the results of logistic regression, which achieved better results.
Results are reported using accuracy and F1 score averaged across the different classes (Macro-F1).

First, we train a generic model on the whole corpus with 5-fold cross-validation. In this case, the model is trained on different topics and dialects.
We show in Table~\ref{tab:results} that, by only adding the {\it country}, {\it topic} and {\it sentiment expression} features directly from the corpus, the performance significantly increases by 13 absolute points.
This indicates the importance of these features for sentiment analysis, and relates back to our manual analysis in which we found sentiment variations across topics and dialects. 

\begin{table*}
\centering
\begin{tabular}{ll|c||c:c|c:c}
\toprule
\multirow{2}*{\bf Features} & & {\bf Generic} & \multicolumn{2}{c|}{\bf Same-Topic} & \multicolumn{2}{c}{\bf Cross-Topic}\\
 & & {\it cross-val} & {\it Politics} & {\it Personal} & {\it Pol-Pers} & {\it Pers-Pol} \\ \midrule
\multirow{2}*{\bf uni/bi-grams} & Acc. & 0.51 & 0.58 & 0.40 & 0.31 & 0.36 \\
 & Macro-F1 & 0.50 & 0.53 & 0.39 & 0.21 & 0.29 \\ \midrule
\multirow{2}*{\bf uni/bi-grams + annotations} & Acc. & 0.63 & 0.64 & 0.56 & 0.47 & 0.50 \\
 & Macro-F1 & 0.63 & 0.62 & 0.54 & 0.37 & 0.44 \\
\bottomrule
\end{tabular}
\caption{Experimental results of a baseline logistic regression model showing the impact of adding the corpus annotation features, and the impact of changing the topic from training to testing.}\label{tab:results}
\end{table*}

We also highlight the impact of change-of-topic between training and testing by conducting two experiments.
In the first experiment, we train our model and test it on data from the same topic, i.e., the {\it topic} feature is implicitly embedded in the model.
In the second experiment, we train our model on data from one topic and test on data from another topic.
We also evaluate, for each experiment, the impact of adding the {\it sentiment expression} feature.
We perform these experiments on the {\it politics} and {\it personal} domains, which are the most frequent topics in our corpus.
We create fixed sets for training and testing with equal sizes in both topics, and use the same splits for all experiments.

Results in Table~\ref{tab:results} show a significant drop in accuracy due to the change-of-topic from training to testing.
This is a typical problem seen when developing cross-domain sentiment models instead of training topic-specific models, which is an expensive solution.
Our corpus allows the development of models for domain adaptation given the availability of topic annotation.
Results also confirm the importance of the {\it sentiment expression} feature, which alone helped improving the performance by more than 10\% absolute.
It can be observed that results on the {\it personal} domain are much lower than those in the {\it politics} domain, which can be attributed to the wider range of sub-topics that can be covered by the {\it personal} domain.
\section{Conclusion}
\label{sec:conclusion}

In this paper, we presented the ArSenTD-LEV; a corpus for sentiment analysis in Arabic Levantine tweets.
Based on a manual analysis that we conducted on a sample of 200 tweets retrieved from the Levant region, we realized the importance of knowing: 1) the topic being discussed by the tweet, 2) the target to which the sentiment was expressed, and 3) the manner the sentiment was expressed, to predict the sentiment of the tweet more accurately.
Consequently, our developed corpus consists of 4,000 tweets collected from Levantine countries (Jordan, Lebanon, Palestine and Syria).
For each tweet, the corpus specifies its overall sentiment, the target to which that sentiment was expressed, and how it was expressed (explicitly or implicitly) and the topic being discussed.
Annotation was performed via crowdsourcing, and annotation guidelines were carefully set to ensure high quality output, which was reflected in the high agreement levels for the different annotated features.

Experimental results confirm the importance of these features.
For instance, including information about the topic and sentiment expression improves the performance of a baseline classifier by more than 10\% absolute.
Furthermore, results confirm the gap that exist between training and testing models on tweets from the same or from different topics.
We also report a significant improvement of 13-14\% when adding the {\it sentiment expression} feature, which suggests some dependency between sentiment polarity and how sentiment is expressed.
It is worth mentioning that for these experiments, we used the manually-annotated features directly from the corpus, which is not a realistic scenario, just to highlight the potential benefits of using these features for sentiment analysis.

Future work include developing accurate machine learning models that leverage the existing annotation to perform both overall and target-based sentiment in Arabic tweets.
It is also interesting, given tweets that are segregated by dialect and topic, to investigate cross-topic and cross-dialect solutions that will mitigate the amount of required resources that will be needed to perform sentiment analysis on any given piece of text.

\section*{Acknowledgment}
This work was made possible by NPRP 6-716-1-138 grant from the Qatar National Research Fund (a member of Qatar Foundation). The statements made herein are solely the responsibility of the authors.

\section{Bibliographical References}
\bibliographystyle{lrec}
\bibliography{references}

\end{document}